# The Optimization of a Novel Prismatic Drive

D. Chablat    S. Caro    É. Bouyer

Institut de Recherche en Communications et Cybernétique de Nantes [1]

UMR CNRS n° 6597, 1 rue de la Noë, 44321 Nantes, France

Damien.Chablat@irccyn.ec-nantes.fr

**Abstract**

The design of a mechanical transmission taking into account the transmitted forces is reported in this paper. This transmission is based on *Slide-o-Cam*, a cam mechanism with multiple rollers mounted on a common translating follower. The design of Slide-o-Cam, a transmission intended to produce a sliding motion from a turning drive, or vice versa, was reported elsewhere. This transmission provides pure-rolling motion, thereby reducing the friction of rack-and-pinions and linear drives. The pressure angle is a relevant performance index for this transmission because it determines the amount of force transmitted to the load vs. that transmitted to the machine frame. To assess the transmission capability of the mechanism, the Hertz formula is introduced to calculate the stresses on the rollers and on the cams. The final transmission is intended to replace the current ball-screws in the *Orthoglide*, a three-DOF parallel robot for the production of translational motions, currently under development for machining applications at *École Centrale de Nantes*.

## 1. Introduction

In robotic and mechatronic applications, whereby motion is controlled using a piece of software, the conversion from rotational to translational motions is usually realized by means of *ball-screws* or *linear actuators*. The both are gaining popularity. However they present some drawbacks. On the one hand, ball-screws comprise a high number of moving parts, their performance depending on the number of balls rolling in the shaft groove. Moreover, they have a low load-carrying capacity, due to the punctual contact between the balls and the groove. On the other hand, linear bearings are composed of roller-bearings to figure out the previous issue, but these devices rely on a form of direct-drive motor, which makes them expensive to produce and maintain.

A novel transmission, called *Slide-o-Cam* is depicted in Fig. (1) and was introduced in [1] to transform a rotational motion to a translational one. Slide-o-Cam is composed of four main elements: (*i*) the frame; (*ii*) the cam; (*iii*) the follower; and (*iv*) the rollers. The input axis on which the cams are mounted, named *camshaft*, is driven at a constant angular velocity by means of an actuator under computer-control. Power is transmitted to the output, the translating follower, which is the roller-carrying slider, by means of pure-rolling contact between the cams and the rollers. The roller comprises two components, the pin and the bearing. The bearing is mounted to one end of the pin, while the other end is press-fit into the roller-carrying slider. Consequently, the contact between the cams and rollers occurs at the outer surface of the bearing. The mechanism uses two conjugate cam-follower pairs, which alternately take over the motion transmission to ensure a positive action; the rollers are thus driven by the cams throughout a complete cycle. Therefore, the main advantages of cam-follower mechanisms with respect to the other transmissions, which transform rotation into translation are: (*i*) the lower friction; (*ii*) the higher stiffness; and (*iii*) the reduction of wear.

---

[1] IRCCyN: UMR n° 6597 CNRS, École Centrale de Nantes, Université de Nantes, École des Mines de Nantes

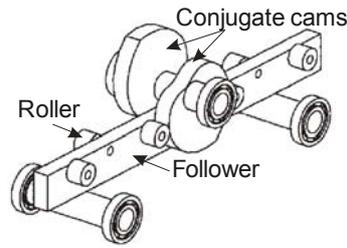

**Figure 1: Layout of Slide-o-Cam**

Many researchers have devoted their efforts to study contact stress distribution and predict surface fatigue life in machine parts under different types of loading. Indeed, when two bodies with curved surfaces, for example, a cam and a roller, are pressed together, the contact is not linear but a surface. The stress occurred may generate failures such as cracks, pits, or flaking in the material. Heinrich Rudolf Hertz (1857-1894) came up with a formula to evaluate the amount of surface deformation when two surfaces (spherical, cylindrical, or planar) are pressed each other under a certain force and within their limit of elasticity.

## 2. Synthesis of Planar Cam Mechanisms

Let the $x$-$y$ frame be fixed to the machine frame and the $u$-$v$ frame be attached to the cam, as depicted in Fig. 2. $O_1$ is the origin of both frames, $O_2$ is the center of the roller, and $C$ is the contact point between the cam and the roller.

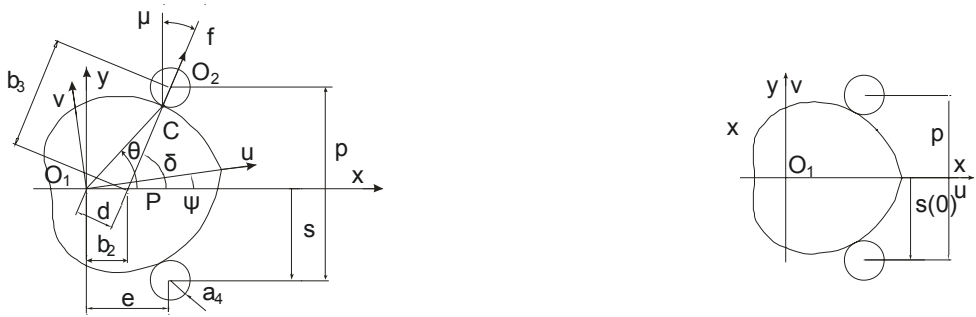

**Figure 2: Parameterization of Slide-o-Cam**   **Figure 3: Home configuration of the mechanism**

The geometric parameters are illustrated in the same figure. The notation used in this figure is based on the general notation introduced in [6], namely, ($i$) $p$ is the pitch, i.e., the distance between the center of two rollers on the same side of the follower; ($ii$) $e$ is the distance between the axis of the cam and the line of centers of the rollers; ($iii$) $a_4$ is the radius of the roller-bearing, i.e., the radius of the roller; ($iv$) $\psi$ is the angle of rotation of the cam, the input of the mechanism; ($v$) $s$ is the position of the center of the roller, i.e, the displacement of the follower, which is the output of the mechanism; ($vi$) $\mu$ is the pressure angle; and ($vii$) **f** is the force transmitted from the cam to the roller.

The above parameters as well as the surface of contact on the cam, are determined by the geometric relations derived from the Aronhold-Kennedy Theorem [2]. As a matter of fact, when the cam makes a complete turn ($\Delta\psi = 2\pi$), the displacement of the roller is equal to $p$, the distance between two rollers on the same side of the roller-carrying slider ($\Delta s = p$). Furthermore, if we consider that Fig. 3 illustrates the home configuration of the roller, the latter is below the $x$-axis when $\psi = 0$. Therefore, $s(0) = -p/2$ and the input-output function $s$ is defined as follows:

$$s(\psi) = \frac{p}{2\pi}\psi - \frac{p}{2} \tag{1}$$

The expressions of the first and second derivatives of $s(\psi)$ are given by:

$$s'(\psi) = p/(2\pi) \quad \text{and} \quad s''(\psi) = 0 \tag{2}$$

The cam profile is determined by the displacement of the contact point $C$ around the cam. The Cartesian coordinates of this point in the $u$-$v$ frame take the form [6]

$$u_c(\psi) = b_2 \cos\psi + (b_3 - a_4)\cos(\delta - \psi) \quad \text{and} \quad v_c(\psi) = -b_2 \sin\psi + (b_3 - a_4)\sin(\delta - \psi) \tag{3}$$

the expression of coefficients $b_2$, $b_3$ and $\delta$ being

$$\begin{aligned}
b_2 &= -s'(\psi)\sin\alpha_1 \\
b_3 &= \sqrt{(e + s'(\psi)\sin\alpha_1)^2 + (s(\psi)\sin\alpha_1)^2} \\
\delta &= \arctan\left(\frac{-s(\psi)\sin\alpha_1}{e + s'(\psi)\sin\alpha_1}\right)
\end{aligned} \tag{4}$$

where $\alpha_1$ is the directed angle between the axis of the cam and the translating direction of the follower. $\alpha_1$ is positive in the counterclockwise (ccw) direction. Considering the orientation adopted for the input angle $\psi$ and the direction defined for the output $s$, as depicted in Fig. 3,

$$\alpha_1 = -\pi/2 \tag{5}$$

The nondimensional design parameter $\eta$ is defined below and will be used extensively in what remains.

$$\eta = e/p \tag{6}$$

The expressions of $b_2$, $b_3$ and $\delta$ can be simplified using Eqs. (1), (2), (4a-c), (5) and (6):

$$\begin{aligned}
b_2 &= \frac{p}{2\pi} \\
b_3 &= \frac{p}{2\pi}\sqrt{(2\pi\eta - 1)^2 + (\psi - \pi)^2} \\
\delta &= \arctan\left(\frac{\psi - \pi}{2\pi\eta - 1}\right)
\end{aligned} \tag{7}$$

From Eq. (7), $\eta$ cannot be equal to $1/(2\pi)$. That is the first constraint on $\eta$. An *extended angle* $\Delta$ was introduced in [7] to know whether the cam profile can be closed or not. Angle $\Delta$ is defined as a root of Eq. (3). In the case of Slide-o-Cam, $\Delta$ is negative, as shown in Fig. 4. Consequently, the cam profile is closed if and only if $\Delta \leq \psi \leq 2\pi - \Delta$.

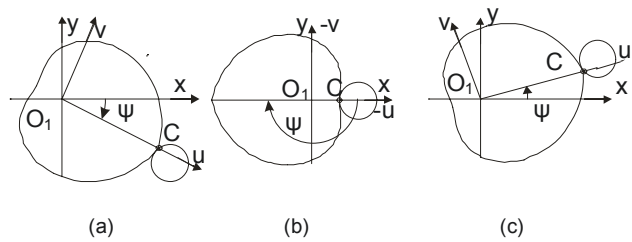

(a)      (b)      (c)

**Figure 4: Orientations of the cam found when $v_c = 0$:**
**(a) $\psi = \Delta$; (b) $\psi = \pi$; and (c) $\psi = 2\pi - \Delta$**

### 2.1. Pitch-Curve Determination

The pitch curve is the trajectory of $O_2$, the center of the roller, distinct from the trajectory of the contact point $C$, which produces the cam profile. $(e, s)$ are the Cartesian coordinates of point $O_2$ in the $x$-$y$ frame, as depicted in Fig. 3. Hence, the Cartesian coordinates of the pitch-curve in the $u$-$v$ frame are

$$u_p(\psi) = e\cos\psi + s(\psi)\sin\psi$$
$$v_p(\psi) = -e\sin\psi + s(\psi)\cos\psi \quad (8)$$

### 2.2. Curvature of the Cam Profile

The curvature of any planar parametric curve, in terms of the Cartesian coordinates $u$ and $v$, and parameterized with parameter $\psi$, is given by [3]:

$$\kappa = \frac{v'(\psi)u''(\psi) - u'(\psi)v''(\psi)}{[u'(\psi)^2 + v'(\psi)^2]^{3/2}} \quad (9)$$

The curvature $\kappa_p$ of the pitch curve is given in [5] as

$$\kappa_p = \frac{2\pi}{p}\frac{[(\psi-\pi)^2 + 2(2\pi\eta-1)(\pi\eta-1)]}{[(\psi-\pi)^2 + (2\pi\eta-1)^2]^{3/2}} \quad (10)$$

provided that the denominator never vanishes for any value of $\psi$, *i.e.*, provided that

$$\eta \neq 1/(2\pi) \quad (11)$$

Let $\rho_c$ and $\rho_p$ be the radii of curvature of both the cam profile and the pitch curve, respectively, and $\kappa_c$ the curvature of the cam profile. Since the curvature is the reciprocal of the radius of curvature, we have $\rho_c = 1/\kappa_c$ and $\rho_p = 1/\kappa_p$. Furthermore, due to the definition of the pitch curve, it is apparent that

$$\rho_p = \rho_c + a_4 \quad (12)$$

Writing Eq. (12) in terms of $\kappa_c$ and $\kappa_p$, we obtain the curvature of the cam profile as

$$\kappa_c = \frac{\kappa_p}{1 - a_4\kappa_p} \quad (13)$$

In [10], the authors assume that the cam profile must be fully convex. The consequence on the design parameters was $\eta > 1/\pi$.

To increase the design parameter space, we accept now to have a non convex cam. But, when the cam push the roller, the sign of the local radius $\rho_c$ must is positive. We study $\rho_c$ for $\eta$ in $]1/(2\pi), 1/\pi]$ and, for $\psi \in ]\Delta, \pi]$, the cam was convex.

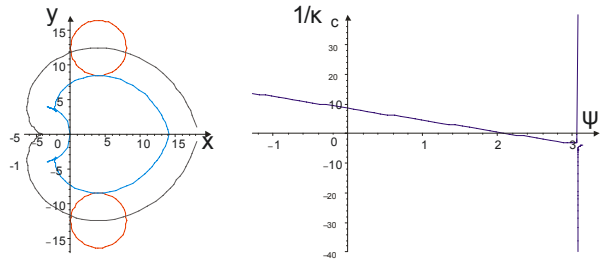

**Figure 5: Cam profile and local curvature of the cam**

In [10], the minimal value of $\rho_c$ is obtain for

$$\psi_{min} = \frac{\pi - \sqrt{4n^2\pi h - n^2 - 4n^2\pi^2 h^2}}{n}$$

Thus when $\rho_c(\psi_{min}) < 0$, the cam profile is not valid. In the former condition is not satisfied, we have a lap-back movement in the curve as depicted in Fig. 5.

### 2.3. Pressure Angle

The pressure angle of cam-roller-follower mechanisms is defined as the angle between the common normal at the cam-roller contact point $C$ and the velocity of $C$ as a point of the follower [3], as depicted in Fig. 2, where the pressure angle is denoted by named $\mu$. This angle plays an important role in cam design. The smaller $|\mu|$, the better the force transmission. In the case of high-speed operations, *i.e.*, angular velocities of cams exceeding 50 rpm, the pressure-angle is recommended to lie advised to be smaller than 30°.

For the case at hand, the expression for the pressure-angle of $\mu$ is given in [3] as:

$$\tan \mu = \frac{s'(\psi) - e}{s(\psi)}$$

Considering the expressions for $s$ and $s'$, and using the parameter $\eta$ given in Eqs.(1), (2)a and (6), respectively, the expression for the pressure angle becomes the pressure angle turns to be:

$$\tan \mu = \frac{n - 2n\pi\eta}{n\psi - \pi}$$

### 2.4. Conjugate Cams

To reduce the pressure angle, several cams can be assembled in the same cam-shaft. We note $m$ the number of cams and $\beta$ the angle of rotation between two adjacent cams, *i.e.*,

$$\beta = \frac{2\pi}{nm} \qquad (14)$$

On the Slide-o-Cam mechanism designed in [1], two conjugate cams with one lobe each and, of which $\beta = \pi$, were used. Figure 6 shows two cam profiles with one and two lobes.

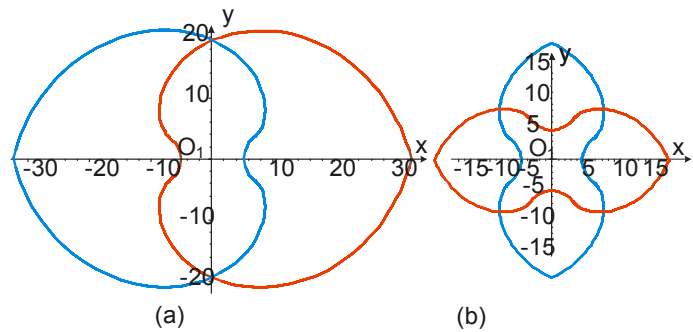

(a)      (b)

**Figure 6: Assembly of two cams with $p = 50$, $a_4 = 10$ and $e = 9$: (a) one lobe; (b) two lobes**

### 2.5. Pressure Angle and Design Parameters

In [10], the authors has studies the influence of design parameters $\eta$, $a_4$ and $n$ on the values of the pressure angle, the cam driving the roller and the influence of the number of conjugate cams on the maximum value of pressure angle. Here is a sum up of the corresponding results::

• **Influence of parameter $\eta$**: *The lower $\eta$, the lower the absolute value of the pressure angle, with $\eta \geq 1/\pi$.*

• **Influence of the radius of the roller $a_4$**: *The lower $a_4$, the lower the absolute value of the pressure angle.*

• **Influence of the number of lobes $n$**: *The lower $n$, the lower the absolute value of the pressure angle.*

• **Influence of the Number of Conjugate Cams**: *The higher the number of conjugate cams, the lower the absolute value of the pressure angle.*

These results are the same if we consider a single cam with several lobes or a two- or three-conjugate-cam mechanism and stay valid for single or conjugate-cam mechanisms and is independent of the number of lobes. However, the contact stress issues and the application were not taken into account in that research work.

### 2.6. Physical constraints

Let us assume that the surfaces of contact are ideal, smooth and dry, with negligible friction. Two relations follow from the strength of materials. Besides, the bearing shafts are subject to shearing, whereas the camshafts are subject to shearing and bending. Consequently, we come up with the following relations:

$$8M_t \left( \frac{2}{\pi \phi_{cam}^3} + \frac{1}{p \phi_{cam}^2} \right) \leq \tau_{c_{max}}$$

$$\frac{8M_t}{p \phi_{bear}^2} \leq \tau_{b_{max}}$$

(15)(16)

where,

is the diameter of the camshaft ($\phi_{cam} = 2(e - a_4)$);

is the diameter of the bearing's shaft ($\phi_{bear} = 2a_4$);

is the torque applied to the camshaft;

is the maximum allowable stress inside the cam axis which cannot be exceeded in the camshaft;

is the maximum stress inside the bearing's axis which cannot be exceeded in the bearing shaft;

### 2.7. Contact Stresses

When two bodies with curved surfaces, for example, a cam and a roller, are pressed together, the contact is not linear but along a surface, due to the inherent material compliance, the stresses developed in the two bodies being three-dimensional. Those contact stresses generate typical failures as cracks, pits, or flaking in the surface material. Heinrich Rudolf Hertz (1857-1894) proposed some formulas to evaluate the width of the band of contact and the maximum pressure, for the case of loaded contact between two cylinders, as depicted in Fig. 7.

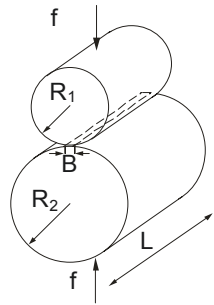

**Figure 7: Parallel cylinders in contact and heavily loaded**

On the one hand, the width $B$ of the band of contact is defined below:

$$B = \sqrt{\frac{16F(K_1 + K_2)R_1 R_2}{L(R_1 + R_2)}}$$

(17)

where,

$F$ is the axial load,

$R_1$ and $R_2$ are the radius of the two cylinders in contact,

$L$ is the width of the cylinders,

$K_1$ and $K_2$ are coefficients which characterize the materials of the two cylinders.

$$K_1 = \frac{1 - \nu_1^2}{\pi E_1}, \quad K_2 = \frac{1 - \nu_2^2}{\pi E_2}$$

(18)

where,

$\nu_1$, $\nu_2$ are the Poisson ratios of the materials of the cylinders 1 and 2,

$E_1$ and $E_2$ are the corresponding Young moduli.

On the other hand, the Hertz pressure is defined as follows:

$$P_h = \frac{4F}{L\pi B} \quad (19)$$

In our case, the two bodies in contact are the roller and the cam. The roller is a cylinder, the cam is not. However, we can approximate the cam locally by a cylinder. Consequently, it is possible to compute the Hertz pressure with respect to $\psi$, which is the angle of rotation of the cam, i.e, the input of the mechanism.

For a given cam profile, the maximum value of the Hertz pressure is obtained for the minimum radius of curvature of the cam. Obviously, the Hertz pressure is a maximum when $B$ is a minimum and $F$, the magnitude of the force $\vec{f}$ transmitted by the cam, is a maximum. Consequently, the higher $B$ and the smaller $F$, the lower the Hertz pressure.

$B$ depends on several parameters, amongst them, the equivalent radius of the contact, $R_{eq}$, which is

$$R_{eq} = \frac{R_1 R_2}{R_1 + R_2} \quad (20)$$

$R_1$ is constant, since it is the radius of the roller, i.e. $R_1 = a_4$ and $R_2$ is the local radius of the cam, i.e. $R_2 = \rho_c$. Therefore, $B$ depends only on $R_2$. Finally, for a given cam profile, $B$ is a minimum when $R_2$ is a minimum. Therefore, to compute the maximum value of the Hertz pressure, we have to consider the lowest value of $R_2$ with respect to $\psi$. Here, we consider the active part of the cam only.

The minimum value of $R_2$, for a two-conjugate cam mechanism, is obtained when $\psi = \pi/n - \Delta$, with $n$ the number of lobes of the cam, and $\Delta$ the extended angle. The load $F$ is computed by

$$F = \sqrt{f_x^2 + f_y^2} \quad (21)$$

where,

$$f_y = \frac{2\pi C_m}{p} = F_0, \quad f_x = \frac{f_y}{\tan\theta} = \frac{F_0}{\tan\theta} \quad (22)$$

with $C_m$ the torque of the motor; $p$, the pitch of the follower; and $\theta$, the angle depicted in Fig. 2. The value of $F$ depends on the mechanism input, $\psi$ whereas $f_x$, attains a minimum when $\psi = \pi/n - \Delta$. Consequently, the Hertz pressure, which is proportional to $F/B$ reaches a maximum when $\psi = \pi/n - \Delta$.

Table 1 presents the maximum Hertz pressure allowed for some common materials. The values are given in MPa, with $P_{stat}$, the allowable pressure for a static load. It is not advised to apply more than 40% of $P_{stat}$ to reach an infinite fatigue life.

**Table 1: Allowable pressures [MPa]**

| Material | $P_{stat}$ maximum | Recommended value of $P_{max}$ |
|---|---|---|
| Stainless steel | 650 | 260 |
| Improved steel | 1600 to 2000 | 640 to 800 |
| Grey cast iron | 400 to 700 | 60 to 280 |
| Aluminum | 62.5 | 25 to 150 |
| Polyamide | 25 | 10 |

Obviously, the maximum allowable pressure depends also on the shape of the different parts in contact. A thin part is less stiff than a thick one. For example, in our case, we can assume that a multilobe cam is less stiff than a single-lobe cam. However, we will only consider here the material for the determination of the allowable pressures values.

## 3. Optimization of a Slide-o-Cam

### 3.1. Influence of the Different Parameters on the Hertz Pressure

The maximum value of the pressure depends on several parameters, namely, the number of conjugate cams, the material of the parts in contact, the geometry of the cam, and the load applied. Therefore, we have different ways to minimize the Hertz pressure.

• Increase the number of conjugate cams, raise the number of conjugate cams, reduce the length of the active part of the cam, and decrease the minimum value of $R_2$. Nevertheless, when we increase the number of conjugate cams, we only consider one contact point. As a matter of fact, we assume that when two cams can drive the rollers, the cam with the smaller absolute value of pressure angle effectively drives the follower;

• Decrease the axial load, which is possible by minimizing the motor torque, or by increasing the pitch. Moreover, since the component of $F$ on the $X$-axis is low when compared with the one of the $Y$-axis, it is more convenient to minimize $f_y$;

• Choose a material with a lower Young modulus, i.e., a more compliant material, thus increasing the surface of contact, hence, decreasing the pressure. However, when the material is more compliant, its plastic domain occurs for smaller stresses.

• Decrease the radius of the cam (defined by $p$ and $e$) and the radius of the roller $a_4$.

### 3.2. Case study

A motivation of this research work is to design a *Slide-o-Cam* transmission for high-speed machines. As mentioned in the introduction, this mechanism should be suitable for the *Orthoglide*, which is a low power machine tool, as shown in Fig. 8 [11]. Here is a list of its features:
• ball screw engine torque $= 1.2$ N.m;
• ball screw engine velocity $= 0$ to $3000$ rpm;
• ball screw pitch $= 20$ mm/turn;
• axial static load $= 376$ N;
• stiffness $= 130$ N/$\mu$m.

We choose here to design Slide-o-Cam with only one lob on each cam and only two cams to reduce the siez of the transmission. Let us assume that the maximum stress that the shafts can support is equal to 150 MPa. The minimum diameter of the bearing shaft $\phi_{bear}$ to transmit the load is equal to 1.8 mm and the cam shaft $\phi_{cam}$ is equal to 3.75 mm when the pitch is equal to 20 mm/turn.

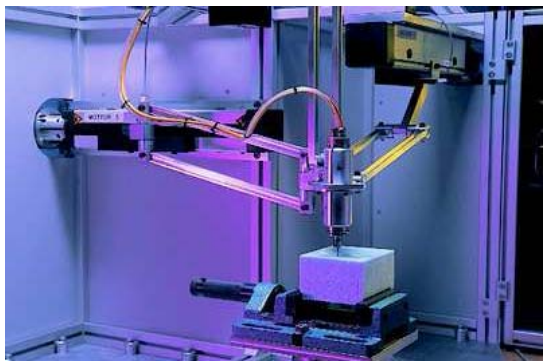

**Figure 8: The Orthoglide (© CNRS Photothèque / CARLSON Leif)**

### 3.3. Optimization of Slide-o-Cam

When the pitch and the torque of the transmission are fixed, four parameters can be changed, $\phi_{cam}$, $\phi_{bear}$, $L$ and the material. Here, we use either steel (respectively aluminum), of which Young modulus is equal to $210\,000$ M$Pa/mm^2$ (respectively $69\,000$ M$Pa/mm^2$). To reduce the size of the transmission, we use only two cams even if the pressure angle can be smaller by using three cams [10].

The pressure angle does not depend on $L$. So, we can represent the isovalues of $\mu$ as a function of $\phi_{bear}$ and $\phi_{cam}$ as depicted in Fig. 9 taking into account the constraint on the maximal value of $\mu$, i.e. $\mu < 30°$ and the curvature of the cam profile.

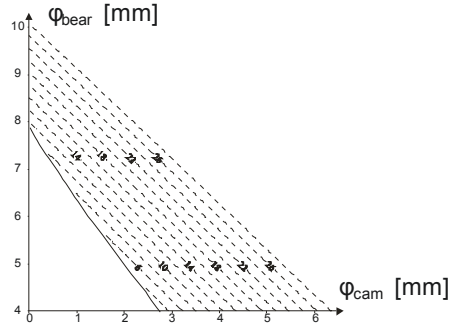

**Figure 9: Isovalues of $\mu_{max}$ for the Orthoglide constraints**

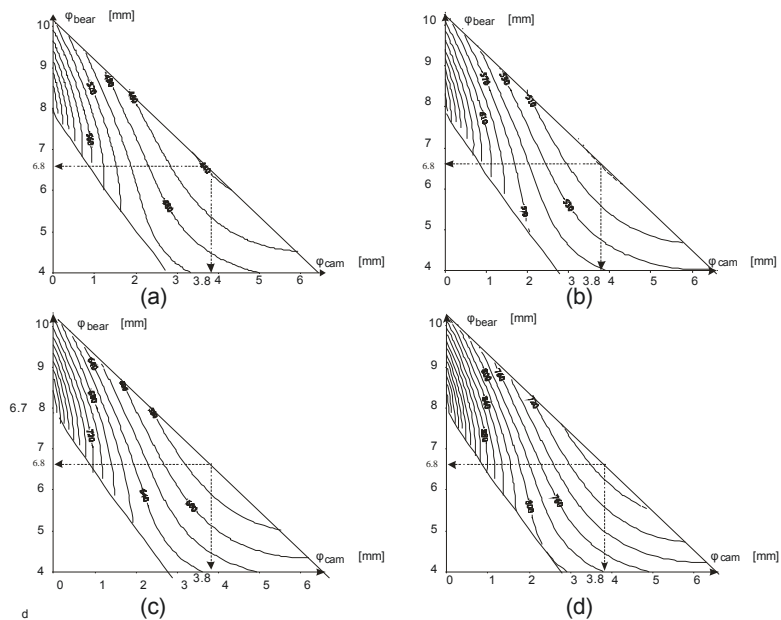

**Figure 10: Isovalues of $P_{h_{max}}$ for (a) $L = 50$ mm, (b) $L = 40$ mm, (c) $L = 30$ mm, (d) $L = 20$ mm**

Due to the mechanical constraints of the Orthoglide, $P_{h_{max}}$ is a minimum for $\phi_{bear} = 6.7$ mm and $\phi_{cam} = 3.8$ mm, which does not depend on $L$ and the material. Figs. 10(a)-(d) depict the isovalues of $P_{h_{max}}$ for $L$ equal 20 mm, 30 mm, 40 mm and 50 mm, respectively. For these values, we have $\tau_c = 145$ MPa and $\tau_b = 5$ MPa. Thus, Fig. 11 plots the variation of $P_h$ as a function of $L$ for steel and aluminum parts for $\phi_{bear} = 6.7$ mm and $\phi_{cam} = 3.8$ mm. In Table 2, the maximal Hertz pressures are smaller for aluminum. However, we have to choose steel parts for the Slide-o-Cam because the advised value for $P_{h_{max}}$ is 150 MPa for aluminum and 800 MPa for steel.

**Table 2: Maximal Hertz pressure for steel and aluminum parts [Mpa]**

| $L$ | Steel | Aluminum |
|---|---|---|
| 10 | 974 | 558 |
| 20 | 689 | 394 |
| 30 | 562 | 322 |
| 40 | 487 | 279 |
| 50 | 435 | 249 |
| 60 | 397 | 228 |

The optimal design with respect to the Hertz pressure is obtained with the maximal value of $L$. However, if we consider the size of the transmission such a parameter has to be bounded. Figure 12 illustrates a possible shape of Slide-o-Cam to replace the ball-screws of the Orthoglide.

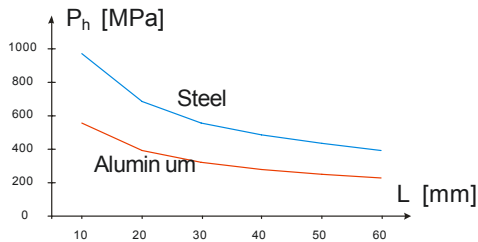

**Figure 11: Variation of $P_{h_{max}}$ as a function of $L$ for $\phi_{bear} = 6.7 mm$ and $\phi_{cam} = 3.8 mm$ for steel and aluminum parts and the Orthoglide constraints**

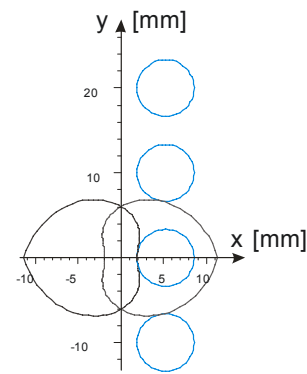

**Figure 12 Slide-o-Cam design for the Orthoglide**

## 4. Summary of results and conclusions

This paper dealt with the optimization of Slide-o-Cam mechanism based on the Pressure angle and Hertz pressure. The kinematic constraints of the Orhtoglide was used to defined to pitch and the input torque but the result was general. For a given input torque and pitch, the optimal design parameters of the cam, i.e. $\phi_{cam}$ and $\phi_{bear}$ are computed when we consider the maximal admissible pressure angle $\mu_{max}$ as a constraint and the strength of the material.

Conversely to previous works, the optimization of the cam parameters based on the Hertz pressure does not converge toward the results obtained by an optimization based on the pressure angle. The optimal values are not the ones obtained for a minimum radius of the roller $a_4$ and a minimum diameter of the cam $\phi_{cam}$. Further research on the fatigue of the cam and the bearing are currently carried on.

## 5. Acknowledgments

This research work was mainly made by Émilie Bouyer during a internship at McGill University from April, the 15th to August, the 31st 2006. The authors would like to acknowledge the Prof. Angeles from McGill University for making the collaborated study possible.

Chablat D., Caro S., et Bouyer E., "The Optimization of a Novel Prismatic Drive", Problems of Mechanics, No 1(26), pp. 32-42, 2007.